\pdfoutput=1

\documentclass[11pt]{article}

\usepackage[final]{coling}

\usepackage{times}
\usepackage{latexsym}
\usepackage{linguex}
\usepackage{comment}
\def\ix.{\ex.\setlength{\Extopsep}{0pt}}

\usepackage[T1]{fontenc}

\usepackage[utf8]{inputenc}

\usepackage{microtype}

\usepackage{inconsolata}

\title{Learning from Impairment: Leveraging Insights from Clinical Linguistics in Language Modelling Research}

\author{Dominique Brunato \\
  Istituto di Linguistica Computazionale "Antonio Zampolli" (CNR-ILC)\\
  ItaliaNLP Lab, Pisa \\
  \texttt{dominique.brunato@ilc.cnr.it} 
  }

\begin{document}
\maketitle
\begin{abstract}
This position paper investigates the potential of integrating insights from language impairment research and its clinical treatment to develop human-inspired learning strategies and evaluation frameworks for language models (LMs). We inspect the theoretical underpinnings underlying some influential linguistically motivated training approaches derived from neurolinguistics and, particularly, aphasiology, aimed at enhancing the recovery and generalization of linguistic skills in aphasia treatment, with a primary focus on those targeting the syntactic domain. We highlight how these insights can inform the design of rigorous assessments for LMs, specifically in their handling of complex syntactic phenomena, as well as their implications for developing  human-like learning strategies, aligning with efforts to create more sustainable and cognitively plausible natural language processing (NLP) models. 
\end{abstract}

\section{Introduction}
The way human language develops and breaks down offers a privileged lens to investigate central aspects of human language competence, including the nuanced sensitivity to sentence acceptability and complexity \cite{Caramazza1978LanguageAA,guijarro2020language,crain1991language,goodman1997inseparability}. This exploration is not only crucial to test linguistic theories on empirical evidence but also has significant implications for crafting language resources informed by cognitive principles. In the current NLP era dominated by Large Language Models (LLMs), such resources are gaining increasing significance \cite{opitz2024natural}, serving as essential benchmarks to unravel the linguistic competence implicitly encoded in neural network representations and possibly shedding light on similarities and differences between how humans and LLMs acquire, represent, and process this knowledge \cite{warstadt2022artificial,10.1162/tacl_a_00254,baroni2022proper}. To date, much research in this direction has been focused on creating targeted diagnostics to assess LLMs' syntactic and semantic abilities inspired by formal linguistics and psycholinguistics  \cite{warstadt-etal-2020-blimp-benchmark, warstadt-etal-2019-neural,ettinger-2020-bert,li-etal-2021-bert,someya-etal-2024-targeted-syntactic}. 
 Alternatively, researchers have drawn inspiration from language acquisition research to explore the use of developmental complexity principles in building more efficient language models \cite{warstadt-etal-2023-findings}, or to compare how humans and algorithms learn language, aiming to identify shared patterns \cite{Evanson2023LanguageAD,yedetore-etal-2023-poor,yedetore-kim-2024-semantic}. 

Building upon this research paradigm that seeks to investigate how human-like language abilities naturally emerge in neural language models--starting with the creation of well-motivated and rigorous test sets and training methods--this position paper   
proposes a novel complementary approach. Specifically, we argue for leveraging the realm of speech and language pathology (SLP) and rehabilitation research as rich data sources to gain insights not only for isolating a set of complexity phenomena but also for devising metrics that facilitate the identification of \textit{hierarchies of complexity} within the same phenomenon, with a particular emphasis on syntactic ones. These nuanced complexity metrics may serve dual purposes. On the one hand, they can be used to craft challenging evaluation protocols to assess the proficiency of deep neural networks in handling key factors underlying sentence complexity. 
On the other one, in line with efforts to develop more sustainable LMs incorporating cognitively plausible mechanisms and trained on reasonably sized data \cite{10.1162/tacl_a_00577,warstadt-etal-2023-findings,huebner-etal-2021-babyberta}, they can inform human-inspired language learning strategies, such as those based on curriculum learning  \cite{bengio2009curriculum,Agrawal2021OnTR,DBLP:journals/corr/abs-2102-03554}, to order training instances based on human-inspired complexity metrics or to design new, linguistically-informed pre-training tasks or targeted prompts aiming at maximizing the generalization of linguistic skills in models.



\section{Linguistic Complexity from the SLP Perspective}

Linguistic complexity plays a crucial role in human sentence processing and also holds strong connections to how both native speakers and deep neural network models perceive the grammaticality of a sentence \cite{linzen-etal-2016-assessing,marvin-linzen-2018-targeted,DBLP:journals/corr/abs-1803-11138}. 

While the long-standing interest for this construct across many communities, defining linguistic complexity is still challenging given its multidimensional and multifarious nature \cite{brunato-etal-2018-sentence}. Common distinctions include `objective vs. agent-related' \cite{dahl_growth_2004} or `absolute vs. relative' \cite{miestamo_grammatical_2008} complexity. Despite different formulations, the first term of these distinctions considers the formal properties of the linguistic system, whereas the second one addresses the issues of cost, difficulty, and level of demand for a language user/learner. From a developmental perspective, 
it is assumed that ``a more complex item is expected to be acquired later than a less complex one'' \cite{di_domenico} or, in different words, that linguistic complexity mirrors ``the order in which linguistic structures emerge and are mastered in second (and, possibly, first) language acquisition'' \cite{doi:10.1177/0267658314536435}. Complexity, in this view, is understood as how an item emerges along stages of development  or the time required for it to be processed in an target-like (i.e. adult or native speaker) manner \cite{Friedmann2021}. 

Like the developmental perspective, the Speech and Language Pathology perspective is user-based but intersects with the specific challenges individuals face due to developmental or acquired language disorders. These challenges can manifest in difficulties with both language comprehension and production, impacting specific or multiple language domains \cite{blackwell1995inducing}. 
Within this broad field, our investigation focuses on aphasiology, specifically agrammatism literature and its clinical implications. Agrammatism is a form of Broca's aphasia, a general linguistic impairment typically resulting from brain damage such as trauma or stroke. It is characterized by structurally impoverished, `telegraphic' speech, with short and fragmented sentences, consisting mainly of content words, while function words and morphemes are frequently omitted. Comprehension of grammatically complex sentences can also be impaired, though understanding of single words and simple sentences is often  intact \cite{goodglass1985agrammatism}. 
 While the key factors shaping the behavioural patterns of loss and preservation in agrammatic aphasia are still debated, it has been suggested that linguistic complexity plays a crucial role in explaining the deficits that these patients have both in production and comprehension \cite{BASTIAANSE200918,avrutin2001linguistics}, as well as in maximizing treatment gains and generalization patterns in language recovery. 







In line with the purpose of this paper, we intend to highlight how the evidence and theoretical underpinnings underlying some influential treatment protocols grounded on linguistic theory can help uncover cognitively-informed \textit{hierarchies of complexity} in language phenomena that can benefit language modelling research in several ways.

\section{Linguistically-Informed Approaches to Language Recovery in Aphasia}

    Developing language treatments that not only enhance trained items but also promote generalization to untrained items/tasks is a key concern in aphasia research. Generalization, defined and applied in various ways over the last half-century, presents challenges for speech-language pathologists in planning, implementing, and measuring treatment protocols \cite{Mayer2024}. Rather than exhaustively reviewing all rehabilitation methods and their efficacy, which is beyond the scope of this paper\footnote{Interested readers can referred to \cite{webster2015time,Mayer2024,fontoura2012rehabilitation}, \textit{inter alia}.}, this section delves deeper on some frameworks rooted in theoretical linguistics \cite{garraffa2020linguistic}, focusing particularly on two of them where linguistic stimuli used in training are grounded in linguistic constructs arranged according to an operationalization of complexity\footnote{Whenever available, all examples are derived from the original stimuli contained in the cited reference papers and presented in Appendix.}. 



\subsection{Mapping Therapy (MT)}
The MT approach moves from the assumption that sentence production and comprehension impairments in agrammatism are due to difficulties in mapping thematic roles (agent, theme) onto their grammatical constituents (subject, object). This can explain the higher failure that these patients exhibit in comprehending and/or producing \textbf{noncanonical sentences}, such as passives and object relative clauses, where the mapping process is indirect compared to active sentences, and particularly in sentences with reversible semantic roles. 
Thus, according to the so-called ``mapping hypothesis'' \cite{schwartz1994mapping}, the proposed treatment program is directed at the remediation of these mapping operations through a metalinguistic training task in which patients are trained to recognize the thematic roles and their syntactic realizations in different structures. The mapping therapy approach has inspired numerous targeted training methods, all emphasizing the essential role of verb argument structure and prompting participants to recognize  thematic role information for correct sentence production. Generally, these protocols follow a stepwise approach with slight variations, organizing training into stages that correspond to increasing levels of complexity in the target stimuli. 
The treatment protocol introduced by \cite{rochon2005mapping} includes four levels (see Appendix \ref{app:MT} for examples). In Level 1 participants are presented with pictures as eliciting material and they are trained to recognize and use only the agent cue to produce active sentences, specifically declaratives and subject clefts. In Level 2, the focus shifts to recognizing and using the theme cue to produce passives and object cleft sentences. As a preparatory step to the final level of treatment, Level 3 introduces the identification of both roles, but in a fixed order. Here, sentences can start with either role, allowing for the elicitation of all four sentence structures. In Level 4, role identification is varied and randomized, with either an agent or theme cue provided.

As in most protocols, learning and generalization effects are usually evaluated across stimuli (using the same structures but different input verbs) and/or across tasks.






\subsection{Treatment of Underlying Forms (TUF)}

TUF is a specific linguistic treatment introduced in \cite{thompson2005treating} that focuses on noncanonical sentence structures and incorporates training for both sentence production and comprehension. It is based on the premise that training underlying, abstract properties of language facilitates generalization to untrained structures with similar linguistic properties. Like MT, it recognizes the key role of verbs and focused training in sentence argument structure. However, it places additional emphasis on the syntactic movement operations (framed in Generative Grammar principles) involved in creating grammatically correct noncanonical sentences. The proposed treatment uses the active form of the expected target sentences to train participants at incremental steps: i.) understand and produce the verb and its arguments in each sentence; ii.) move the proper sentence constituents to form target sentence structures; iii.) produce the surface form of the target sentence; iv) comprehend and produce the verbs and verb arguments in their noncanonical position. What sets TUF apart from other protocols based on explicit learning is its proposal that training more complex structures first yields better outcomes than the traditional method of gradually increasing item complexity. This approach has been systematized in the\textbf{ Complexity Account of Treatment Efficacy (CATE)}, according to which ``training complex structures results in generalization to less complex structures when untreated structures encompass processes
relevant to (i.e., are in a subset relation to treated ones'' \cite{Thompson-CATE}.
As mentioned, the construct of syntactic complexity relates to sentences containing a type of unbounded dependencies, i.e. constructions whose correct understanding/processing requires the computation of grammatical relationship between phrases that are pronounced in a position different from the one where they are interpreted. 
These constructions, also commonly named ``filler-gap'' dependencies in psycholinguistics, are known to pose significant challenges for human language processing, including in non-impaired speakers. The challenge is particularly pronounced when the ``filler'' (i.e., the pronounced element) is far from the ``gap'' (the position in which the element is interpreted). CATE specifically targets noncanonical sentences derived from \textit{wh-movement}, particularly focusing on \textbf{object-extracted dependencies} in three  forms: object wh-questions (WH), object clefts (OC), and object relative clauses (OR) (see Appendix \ref{app:CATE} for  examples). It assumes that a hierarchy of increasing complexity can be traced from the first to last structure, which can be explained according to movement operations and resulting parse tree depth (for details, see the reference papers). 

 The validity of this hierarchy is supported by the patterns of loss observed in agrammatic comprehension, especially in the effectiveness of predicted recovery trajectories. It has been observed that treatment targeting object relative clauses successfully generalized to untreated object clefts and object wh-questions, but not the other way around. For the purpose of this paper, we do not aim to argue the clinical effectiveness of CATE compared to other treatments that follow a more intuitive approach, i.e., progressing from simpler to more complex training examples. Instead, we aim to use the findings within this framework as evidence for the existence of syntactic structures that can be systematically ordered by complexity.


\subsection{Syntax Stimulation Program (SSP)}

 This method, also known as Helm Elicited Language Program for Syntax Stimulation (HELPSS) \cite{helm1981helm}, aims to enlarge the repertoire of grammatical structures by agrammatic aphasic patients.  Differently from the previously described protocols, it  focuses on practicing the surface form of target structure rather then training the underlying structure and computations. The training protocol is sequentially arranged from repetition to spontaneous production in context. The program targets eleven types of sentences, which are ordered for level of difficulty as follows: Imperative Intransitive; Imperative Transitive; Wh- Interrogative; Declarative Transitive; Declarative Intransitive; Comparative; Passives; Yes-No Questions; Direct-Indirect Object; Embedded Sentences; Future. Each typology is trained within a story completion
format comprising two levels of difficulty: the first uses delayed repetition of the expected target phrase, the second removes the benefit of repetition from the patient.




\section{Insights from Aphasia Treatment for Language Modelling Research}

Our proposal suggests that training protocols for aphasia treatment can play a crucial role in developing more robust and cognitively plausible LMs informed by human language (re-)learning.  Specifically, we outline contributions in two main areas covering both training and evaluation frameworks.

\subsection{Evaluation via Complexity Hierarchies}
So far, a large variety of benchmarks informed by theoretical and experimental linguistics has been introduced to test the linguistic competencies of deep neural language models \cite{warstadt-etal-2019-neural,warstadt-etal-2020-blimp-benchmark,hu-etal-2020-systematic}. However, none has tested the model sensitivity to the same phenomenon (e.g. Wh-movement) arranging stimuli at incremental levels of complexity, as proposed e.g. by the CATE approach. This involves not only manipulating for grammaticality sentences hierarchically arranged at different complexity levels, and testing the model's response, but also assessing whether these sentences influence the model's ability to perform high-level reasoning tasks, such as natural language inference.

\subsection{Learning Strategies} 

\noindent\textbf{Curriculum Learning with Human-Inspired Complexity}: Ranking training data based on complexity hierarchies derived from neurolinguistics and clinical treatment could be a viable strategy to optimize the curriculum for LMs, mimicking how humans learn progressively from simpler to more complex structures. Such hierarchies can also serve as benchmarks for evaluating the effectiveness of other training data optimization methods, including those inspired by language acquisition scenarios. 
While studies so far have reported mixed or negative findings on developmental-inspired learning strategies for LM training, the prevailing assumption is that child-directed speech provides an optimal foundation to foster linguistic generalization capabilities in smaller-scale training regimes \cite{huebner-etal-2021-babyberta,martinez-etal-2023-climb,DBLP:journals/corr/abs-2305-07759}. However, a SLP-informed curriculum presents an alternative, data-efficient pretraining technique by systematically organizing training data around explicit syntactic complexity metrics. 
Unlike the variability inherent in child-directed speech, this approach creates a structured pathway through progressively complex syntactic forms, which could accelerate generalization and adaptability to novel linguistic constructions.

\noindent \textbf{Linguistic Learning Objectives}: Aphasia treatment protocols can inform the creation of new pre-training  objectives for LMs. These objectives, recast based on linguistic tasks proposed by clinicians, can supplement traditional language modelling approaches by injecting a structured linguistic bias at a foundational level. Notably, while linguistically-enhanced pre-training tasks like semantic role labeling have shown promise  \cite{Cui2022LERTAL,zhou-etal-2020-limit}, their effectiveness also combined with training on progressively more complex sentences in terms of diverse argument structures, as in the MT protocol, remains unexplored.

\noindent \textbf{Targeted Prompts for Improved Generalization}: Insights from aphasia research can guide the creation of targeted prompts for use in knowledge distillation scenarios. These prompts, designed to address specific linguistic constructs, improve the model's capability to generalize and master complex structures. Similar to aphasia therapy, where patients progress through increasingly complex linguistic tasks to (re-)build language abilities, this approach can be applied in LLMs. Structuring prompts to begin with simpler forms and gradually introduce more complex sentence constructions may enable the student model to effectively learn a wide range of linguistic phenomena.

\section*{Limitation and Future Directions}
As an opinion paper, our work proposes a framework for language modelling research without providing extensive empirical evidence through comparisons with other training strategies or evaluation settings for language models. We have not discussed possible implementation strategies in detail, but potential approaches could include starting from available examples in the literature to identify relevant templates and using synthetic data inspired by them to augment data. Alternatively, sentence constructions such as those listed in the SSP approach might also be searched for in available treebanks for different languages. Notably, while we focus on examples from the original papers (often in English), adaptations exist for many other languages. 
Moreover, our primary focus was on linguistically-based training methods developed in the field of aphasia, which involve explicit teaching of grammatical structures or syntactic rules to promote generalization and foster metalinguistic knowledge primarily through textual input, often supported by picture only. However, these approaches, while significant, are not unique within the field of aphasia rehabilitation and can be enhanced by incorporating other modalities input alongside the linguistic one. Integrating such multimodal training objectives into language models is a promising avenue for future research. 

\section*{Acknowledgments}
The author gratefully acknowledges the support of the project ``LODE – The
Language Of Dreams: the relationship between sleep mentation, neurophysiology, and neurological disorders '' - PRIN 2022 2022BNE97C\_SH4\_PRIN2022 funded by the Italian Ministry of University and Research, and of the project XAI-CARE - PNRR-MAD-2022-12376692 under the NRRP MUR program funded by the NextGenerationEU.

\bibliography{custom}
\appendix

\section{Examples from the Mapping Therapy Protocol \cite{rochon2005mapping}} \label{app:MT}

\textbf{Level 1, Agent Role Cue}: Target sentences are reversible, 2-argument structures in the form of actives and subject clefts.

 Target sentence: ``The nurse chases the tall teacher''
 
 Examiner says:
\begin{itemize}
\item This is a picture about chasing.
\item The verb in the sentence is `chases'.
\item In this picture the one being chased is the tall teacher (Theme).
\item The one doing the chasing is the nurse (Agent)
\item Please make a sentence starting with the nurse\textsuperscript{*}.
\end{itemize}

\textbf{Level 2, Theme Role Cue}: Target sentences are reversible, 2-argument structures in the form of passives and object clefts.

Target sentence: “The farmer was hugged by the soldier”.

Examiner says:
\begin{itemize}
\item This is a picture about hugging.
\item The verb in the sentence is “hugged”.
\item In this picture the one being hugged is the farmer.
\item The one doing the hugging is the soldier.
\item Please make a sentence starting with the farmer.
\end{itemize}

\textbf{Level 3, Agent and Theme Role Cue}: Target sentences are reversible, 2-argument structures in all of the four possible forms:  actives, subject clefts, passives, object clefts.

Target sentence: “The baker was called by the judge”.

Examiner says:
\begin{itemize}
\item This is a picture about calling.
\item The verb in the sentence is “called”.
\item In this picture the one doing the calling is the judge.
\item The one being called is the baker.
\item Please make a sentence starting with the baker.
\end{itemize}

\textbf{Level 4, Agent or Theme Role Cue}: Target sentences are reversible, 2-argument structures in all the four possible forms:  actives, subject clefts, passives, object clefts.

Target sentence: “It is the author that is shooting the farmer”.

Examiner says:
\begin{itemize}
\item This is a picture about shooting.
\item The verb in the sentence is “shooting”.
\item In this picture the one doing the shooting is the author.
\item The other person is the farmer.
\item Please make a sentence starting with the author.
\item Please start your sentence with ``It is''.

\end{itemize}

* At all levels, when the target was a subject cleft or object cleft sentence, the examiner added: “Please begin the sentence starting with ‘it is’.”

\section{Examples from the Complexity Account of Treatment Efficacy (CATE) \cite{Thompson-CATE}}\label{app:CATE}

\ex.
    \a. (OR) The man saw the thief who the artist chased \_.
     \b. (OC) It was the thief who the artist chased \_. 
    \c.  (WH) Who did the artist chase \_?

\ex.
    \a. (OR) The man saw the husband who the wife covered \_.
     \b. (OC) It was the husband who the wife covered \_. 
    \c.  (WH) Who did the wife cover \_?  

\ex.
    \a. (OR) The man saw the guest who the waiter watched \_.
     \b. (OC) It was the guest who the waiter watched. \_. 
    \c.  (WH) Who did the waiter watch\_?



\end{document}